\documentclass[sigconf, table]{acmart}

\usepackage{array}
\usepackage{graphicx}
\usepackage{listings}
\usepackage{booktabs} 
\usepackage{multirow} 
\usepackage{subcaption}
\usepackage{verbatim} 
\usepackage{mathtools}
\usepackage[linesnumbered,ruled,vlined]{algorithm2e}
\usepackage{color}
\usepackage{url}

\usepackage{flushend}

\usepackage{soul}

\usepackage{caption} 
\mathtoolsset{firstline-afterskip=0pt}

\colorlet{dark-green}{green!70!black}

\newcommand{\jf}[1]{\textcolor{orange}{#1}}

\newcolumntype{A}{>{\raggedright\arraybackslash}m{1.65cm}}
\newcolumntype{B}{>{\raggedright\arraybackslash}m{1.75cm}}

\setcopyright{acmlicensed}

\acmDOI{}

\acmISBN{978-x-xxxx-xxxx-x/YY/MM}

\acmConference{GECCO '18}{July 15-19, 2018}{Kyoto, Japan}
\acmYear{2018} 
\copyrightyear{2018}

\acmPrice{15.00}

\begin{document}
	
	\title{Analysing Symbolic Regression Benchmarks under a Meta-Learning Approach} 
    
	
	\author{Luiz Otavio V. B. Oliveira, Joao Francisco B. S. Martins, Luis F. Miranda, Gisele L. Pappa}
	\affiliation{%
		\institution{Universidade Federal de Minas Gerais, Department of Computer Science}
		\city{Belo Horizonte} 
		\country{Brazil}
	}
	\email{[luizvbo, joaofbsm, luisfmiranda, glpappa]@dcc.ufmg.br}
	
	\renewcommand{\shortauthors}{Oliveira et al.}
	
	\begin{abstract}
		
  The definition of a concise and effective testbed for Genetic Programming (GP) is a recurrent matter in the research community. This paper takes a new step in this direction, proposing a different approach to measure the quality of the symbolic regression benchmarks quantitatively. The proposed approach is based on meta-learning and uses a set of dataset meta-features---such as the number of examples or output skewness---to describe the datasets. Our idea is to correlate these meta-features with the errors obtained by a GP method. These meta-features define a space of benchmarks that should, ideally, have datasets (points) covering different regions of the space. An initial analysis of 63 datasets showed that current benchmarks are concentrated in a small region of this benchmark space. We also found out that number of instances and output skewness are the most relevant meta-features to GP output error. Both conclusions can help define which datasets should compose an effective testbed for symbolic regression methods.

	\end{abstract}

    \copyrightyear{2018}
    \acmYear{2018}
    \setcopyright{acmlicensed}
    \acmConference[GECCO '18 Companion]{Genetic and Evolutionary Computation Conference Companion}{July 15--19, 2018}{Kyoto, Japan}
    \acmPrice{15.00}
    \acmDOI{10.1145/3205651.3208293}
    \acmISBN{978-1-4503-5764-7/18/07}

	%
	%

	
	
	
	\maketitle
	
	\section{Introduction}

The quest for better Genetic Programming (GP) algorithms---e.g., capable of overcoming known drawbacks, presenting new interesting properties or operating with fewer resources---is as important as the search for better ways of evaluating these methods and comparing them under different aspects. It is necessary to consistently and efficiently identify the scenarios where the new algorithm excels and how it compares to its predecessors. Otherwise, we may reach the wrong conclusions, which may mislead future studies and indicate an inappropriate method to solve the problem at hand.

The definition of a concise and effective testbed for GP is a recurrent matter in the research community. In the 14th Genetic and Evolutionary Computation Conference, \citet{mcdermott2012genetic} brought to light the need for a redefinition of the datasets employed by the GP community. This work resulted in an extended study \cite{white2013better}, which listed possible good and bad choices for benchmarking in GP. However, the quality of the benchmarks was measured subjectively using the feedback obtained from a GP community survey and the 14th GECCO attendees.

This work takes a different direction by analysing the datasets employed by the GP community using meta-features captured from the data. Our main idea is to give directions towards the proposal of an approach to measure the quality of the benchmarks quantitatively. Although previous works have already analysed GP datasets under a quantitative perspective \cite{nicolau2015guidelines,dick2015reexamination}, to the best of our knowledge this is the first work to use meta-features. The objective of this paper is to characterize a set of GP benchmarks in symbolic regression according to the relationship of meta-features extracted from them and the error generated by GP.

In order to do that, we first identified the set of papers from GECCO 2013 to 2017 that dealt with symbolic regression. These papers used a set of 80 datasets. For 63 of them, we extracted a set of 11 meta-features, including, for instance, output skewness and the number of attributes, and ran a genetic programming (GP) method to obtain the normalized root mean square error (NRMSE). We then analysed the distribution of the values of these meta-features across the datasets according to the NRMSE obtained by GP, and plotted the ``space of benchmarks'' defined by these meta-features. The idea is to have dataset benchmarks that cover different regions of this space in order to assess whether new methods are really effective to solve the symbolic regression problem. The results showed that most of the datasets used as benchmarks in the literature present very similar meta-features and are concentrated in a small region of this benchmark space, and that output skewness and number of instances are the most relevant meta-features to predict the GP error.

The remainder of this paper is organized as follows. Section \ref{sec:rw} discusses different approaches proposed to objectively assess the performance of learning methods, Section \ref{sec:methodoly} presents the methodology adopted to characterize the datasets, while Section \ref{sec:experimental_analysis} analyses the experimental procedures conducted and their results. Finally, Section \ref{sec:conclusions} draws conclusions about our findings and outlines future work directions.

	\section{Related Work}\label{sec:rw}

As previously mentioned, the works in \cite{white2013better,mcdermott2012genetic} propose the first initiative to perform a critical analysis of the datasets used to benchmark genetic programming in general, using a somehow qualitative approach to define what a good benchmark is. Interestingly to say, many of the datasets suggested as blacklisted in \cite{white2013better} (e.g., quartic and lower order polynomials) were used in 14 out of 26 papers identified as dealing with symbolic regression in GECCO papers from 2013 to 2017.

Following these initial works, 
\citet{nicolau2015guidelines} analysed symbolic regression datasets presented by \citet{mcdermott2012genetic}, comparing different aspects of these benchmarks. They focused on synthetically generated datasets, 
exploring the sampling strategy and size and the addition of artificial noise to the data. The authors compared GP and Grammatical Evolution (GE) to two particularly simple baselines---a constant, corresponding to the average response observed in the training set, and a linear regression model. Results highlight a correlation between the distribution of the response variable and the performance of evolutionary algorithms. For datasets with a response variable with high variance, models induced by linear regression or by constant values were as good as or better than the ones induced by GP and GE.

In this same direction,
\citet{dick2015reexamination} presented a quantitative analysis of the Human oral bioavailability dataset, used to guide the development of different GP-based methods. The authors analyse features
with the assistance of information gain and Random Forests feature selection procedures, sampling variance and the performance of GP-based methods with Lasso, $k$-Nearest Neighbour and Random Forests for regression. The analysis shows that the dataset presents flaws that indicate it should be avoided in future works.


Although not applied to regression problems, Mu{\~n}oz and colleagues \cite{munoz2017generating,munoz2018instance} present a framework to build a two-dimensional space based on meta-features of classification problems to identify regions where classification algorithms excel---and fail---and to generate datasets with the aim of enriching the diversity of classification problems in the meta-feature space.


Following Mu{\~n}oz and colleagues \cite{munoz2017generating,munoz2018instance}, the work presented here also explores the idea that the performance of algorithms can be visualized and pockets of the instance space corresponding to algorithm strengths and weaknesses can be identified. We extend that idea, however, to regression problems tackled by GP-based methods.

	\section{Methodology}
\label{sec:methodoly}

As previously explained, the main objective of this paper is to characterize a set of GP benchmarks in symbolic regression according to the relationship of their features and the error generated by the GP.
Recall that a symbolic regression task consists of inducing a model that maps inputs to outputs. More precisely, given a finite set of input-output pairs representing the training instances $T=\{(\mathbf{x}_i,y_i)\}_{i=1}^n$---with $(\mathbf{x}_i,y_i)\in \mathbb{R}^d\times \mathbb{R}$ and $\mathbf{x}_i=[x_{i1}, x_{i2},\ldots, x_{id}]$, for $i=\{1,2,\ldots,n\}$---we define $X=[\mathbf{x}_1,\mathbf{x}_2,\ldots,\mathbf{x}_n ]^T$ and $Y=[y_1,y_2,\ldots,y_n]^T$ as the matrix $n\times d$ of inputs and the $n$-element output vector, respectively. A symbolic regression consists then in inducing a model $p: X \rightarrow Y$ such that $\forall(\mathbf{x}_i,y_i) \in T: p(\mathbf{x}_i) = y_i$.

In order to characterize datasets used to benchmark methods to solve symbolic regression problems, we performed the following tasks:

\begin{enumerate}
\item We looked at the datasets that have been used to evaluate GP methods at GECCO papers from 2013 to 2017.
\item We extracted from these datasets a set of meta-features in order to characterize them. It is important to mention that the area of meta-learning has extensively studied attributes to characterize datasets, and this is still an open problem \cite{brazdil2008metalearning}. We have chosen 5 meta-features and generated them for both the training and test sets. In addition, we also considered the number of features.
\item We ran a canonical GP method in the selected datasets and recorded the normalized root mean squared error (NRMSE) for each dataset.
\item We created a meta-dataset, where each feature represents a meta-feature and is associated with the NRMSE generated for that respective dataset.
\item We used this dataset to determine which features were more related to the NRMSE obtained.
\end{enumerate}

The next sections detail each of these steps.

\subsection{Datasets}

We started by compiling a list of 80 datasets employed in 26 papers working with symbolic regression problems published at GECCO from 2013 to 2017. From these 80 datasets, we could not find the description of two synthetic datasets---Sext \cite{krawiec2014behavioral} and Nguyen-12 \cite{krawiec2014behavioral,liskowski2017discovery}---and we could not find seven real-world datasets available on line---Dow chemical \cite{nicolau2016managing}, Plasma protein binding level (PPB) \cite{castelli2015geometric,oliveira2016dispersion,goncalves2017unsure}, Tower data \cite{lacava2015genetic,lacava2016epsilon,oliveira2016dispersion}, NOX  \cite{arnaldo2014multiple,arnaldo2015building}, Wind (WND) \cite{lacava2016epsilon}, Median lethal dose (toxicity/LD50) \cite{castelli2015geometric,goncalves2017unsure} and Human oral bioavailability  \cite{castelli2015geometric,dick2015reexamination,oliveira2016dispersion,goncalves2017unsure}\footnote{We intended to analyze only datasets freely available for download, in order to make the access to the data easier for the reader.}. 

In addition, we decided not to use the \textit{million song dataset}, given time restrictions, and Korns-2, Korns-3, Korns-5, Korns-6, Korns-8, Korns-9, Korns-10, given inconsistencies on the generated data---these datasets generate inputs that lead to inconsistent data, caused by division by zero, logarithm of zero or negative number and square root of negative number---ending up with 63 datasets. Tables~\ref{table:real_datasets} and \ref{table:synth_datasets} describe the real and the synthetic datasets, respectively.
For the synthetic datasets, the training and test sets are sampled independently, according to two strategies. $U[a, b, c]$ indicates a uniform random sample of size $c$ drawn from the interval $[a, b]$ and $E[a, b, c]$ indicates a grid of points evenly spaced with an interval $c$, from $a$ to $b$, inclusive.

\begin{table*}[t]
	\rowcolors{2}{gray!25}{white}
	\caption{Real datasets used in papers from GECCO 2013 to GECCO 2017.}
	\begin{center}
		\begin{tabular}{llrrr}
			\toprule
			\textbf{Abbreviation} & \textbf{Dataset} & \textbf{\# of features} & \textbf{\# of instances} & \textbf{Source}\\
			\cmidrule(r{0em}){1-5}
			ABA & Abalone\textsuperscript{1} & 8 & 500 & \cite{thuong2017combining}\\
			AFN & Airfoil self-noise & 6 & 1503 & \cite{oliveira2016dispersion}\\
			BOH & Boston housing & 14 & 506 & \cite{whigham2015examining,dick2015reexamination,lacava2016epsilon,thuong2017combining}\\
			CCP & Combined cycle power plant & 4 & 9568 & \cite{medernach2016new}\\
			CPU & Computer hardware & 9 & 209 & \cite{oliveira2016dispersion}\\
			CST & Concrete strength & 9 & 1030 & \cite{medernach2016new,oliveira2016dispersion}\\
			ENC & Energy efficiency, cooling load & 9 & 768 & \cite{arnaldo2014multiple,arnaldo2015building,lacava2016epsilon,oliveira2016dispersion}\\
			ENH & Energy efficiency, heating load & 9 & 768 & \cite{arnaldo2014multiple,arnaldo2015building,lacava2016epsilon,oliveira2016dispersion}\\      
			FFR & Forest fires & 13 & 517 & \cite{oliveira2016dispersion}\\
			MSD & Million song dataset\textsuperscript{2} & 90 & 1000000 & \cite{arnaldo2015building}\\
			OZO & Ozone\textsuperscript{3} & 73 & 2536 & \cite{thuong2017combining} \\
			WIR & Wine quality, red wine & 12 & 1599 & \cite{arnaldo2014multiple,arnaldo2015building,oliveira2016dispersion}\\
			WIW & Wine quality, white wine & 12 & 4898 & \cite{arnaldo2014multiple,arnaldo2015building,oliveira2016dispersion}\\
			YAC & Yacht hydrodynamics & 7 & 308 & \cite{medernach2016new}\\
			\midrule[\heavyrulewidth] 
		\end{tabular}
	\end{center}
	\textsuperscript{1} 500 randomly selected instances. The feature \textit{sex} was represented as dummy variable. \\
    \textsuperscript{2} Dataset was not used due to execution time  restrictions. \\
    \textsuperscript{3} Missing values were replaced by the feature mean. \\
	\label{table:real_datasets}
\end{table*}

\subsection{Meta-dataset predictive features}

A set of six dataset-related meta-features were chosen, most of them with training and test equivalents, with exception of the number of features, which remains the same in all cases. They were:

\begin{enumerate}
\item Number of features;
\item Number of instances; 
\item Target feature (output) skewness, defined in Eq.~\ref{eq:skw}, where $n$ is the number of instances, and $\bar{x}$ is the average of the sample;

\begin{equation} \label{eq:skw}
Skewness = \frac{n\sqrt{n-1}}{n-2}\cdot\frac{\sum_{i=1}^{n}(x_i-\bar{x})^3}{[\sum_{i=1}^{n}(x_i-\bar{x})^2]^{3/2}}
\end{equation}
\item Standard deviation of the target feature;
\item Mean of the absolute feature-target correlation, calculated as the average of the Spearman correlation among each feature and the target output. The higher the value of this feature, the simpler the dataset;
\item Linearity measure: In order to evaluate the ``linearity'' of a dataset, we used the coefficient of determination ($R^2$) of the model induced by a linear regression on the dataset. Recall that $R^2$ determines the percentage of variation in the output variable that can be explained by the linear relationship between the predictive features and the output. A low $R^2$ means there is not a strong linear relationship between predictive and output features.
\end{enumerate}

We selected meta-features easily computed and directly related to the regression task. The number of instances reflect the available data for fitting and model evaluation; the number of features affect the input space; the  skewness and standard deviation of the output capture the distribution of the target feature; the correlation feature-target captures the relation between the input and output features; and the linearity of the data is related to the shape of the function that generated the data.

It is important to mention that it makes sense to look at feature in both the training and test sets, as
we are analyzing the quality of the datasets themselves, and not their generalization ability. In a future work, we intend to analyze the capability of modeling GP performance according to the meta-features.


\subsection{Meta-dataset output feature}

After extracting the meta-features from the 63 datasets, we ended up with a meta-dataset of 63 instances, each described by 11 features. The next step was to associate a metric of error to each of these instances. We adopted the Normalized Root Mean-Squared Error (NRMSE) \cite{miranda2017how} obtained by $m$ runs of a GP as the meta-dataset output.

The NRMSE was chosen because, as we want to understand the relationship between the GP performance across different datasets, a normalized performance metric is more appropriate. We adopted the median over different executions of the GP method given by:

\begin{equation}\label{eq:nrmse}
	NRMSE=\frac{RMSE\cdot\sqrt{\frac{n}{n-1}}}{\sigma_Y}=
	\sqrt{\frac{\sum\limits_{i=1}^n{(y_i-f(x_i))^2}}{\sum\limits_{i=1}^n{(y_i-\bar{Y})^2}}}\enspace,
\end{equation}

\noindent
where $\bar{Y}$ and $\sigma_Y$ are, respectively, the mean and standard deviation of the vector $Y$, composed by the expected outputs given by the training (or test) set, and $f$ is the model (function) induced by the GP.

We defined different strategies for the GP experiments according to the nature and source of the datasets. For real datasets, we randomly partitioned the data into five disjoint sets of the same size and carried out the experiments six times with a 5-fold cross-validation (6 $\times$ 5-CV). For the synthetic datasets, the data was sampled only once and the experiments were repeated 30 times with the same data.

We adopted the GP implementation from the \textit{gplearn} Python package \cite{stephens2018gplearn}. The fitness was defined as the NRMSE and the function set composed by $\{+,-,\times, AQ, sqrt, sin \}$, where the Analytic Quotient (AQ) \cite{ni2013use} replaces the (protected) division and is defined as:

\begin{equation}
	\label{eq:aq}
	\textit{AQ}(a,b)=\frac{a}{\sqrt{1+b^2}}\enspace.
\end{equation}

\subsection{Meta-features Analyses}

Having the meta-dataset defined, our final step was to perform a series of analyses to better understand the relations between the meta-features and the NRMSE.
They were:

\begin{enumerate}
	\item The meta-features relevance to determine the NRMSE, obtained according to a Random Forest Regressor (RFR). We adopted the implementation from Scikit-learn \cite{pedregosa2011scikit} to determine the importance of the features.
\item The meta-models generated using the meta-dataset created to predict the NRMSE of a GP when using a dataset with similar features. We adopted two approaches:
	\begin{enumerate}
    	\item We fitted a linear regression model to the meta-dataset, using the NRMSE as the response variable. 
        \item We captured the variance of the meta-features by reducing the meta-dataset to two features, represented by the two principal components generated by the Principal Component Analysis (PCA) method \cite{hotelling1933pca}---applied after the meta-features were normalized. Then we fitted a linear regression to this two-dimensional data, using the NRMSE as the response variable. This strategy allows us to visualize the meta-features (the components representing them), along with the NRMSE, in a three-dimensional plot.
    \end{enumerate}
\end{enumerate}






	\section{Experimental Analysis}
\label{sec:experimental_analysis}

This section shows the results of the analyses described in the previous section\footnote{The code used in our experimental analysis is freely available for download on GitHub at \url{https://github.com/laic-ufmg/gp-benchmarks}}. Note that the output values were obtained by running a canonical GP framework with a population of 1,000 individuals evolved for 50 generations. Subtree crossover, subtree mutation, hoist mutation and point mutation were applied with probabilities 0.85, 0.05, 0.05 and 0.05, respectively, and the tournament selection had size 10.
All other parameters were set to their default value.

\subsection{Meta-dataset analysis}

\begin{figure*}[ht]
	\captionsetup[sub]{skip=0mm}
	\captionsetup{skip=1mm}
    \begin{subfigure}[ht]{0.33\textwidth}
    	\centering
		\includegraphics[width=\linewidth]{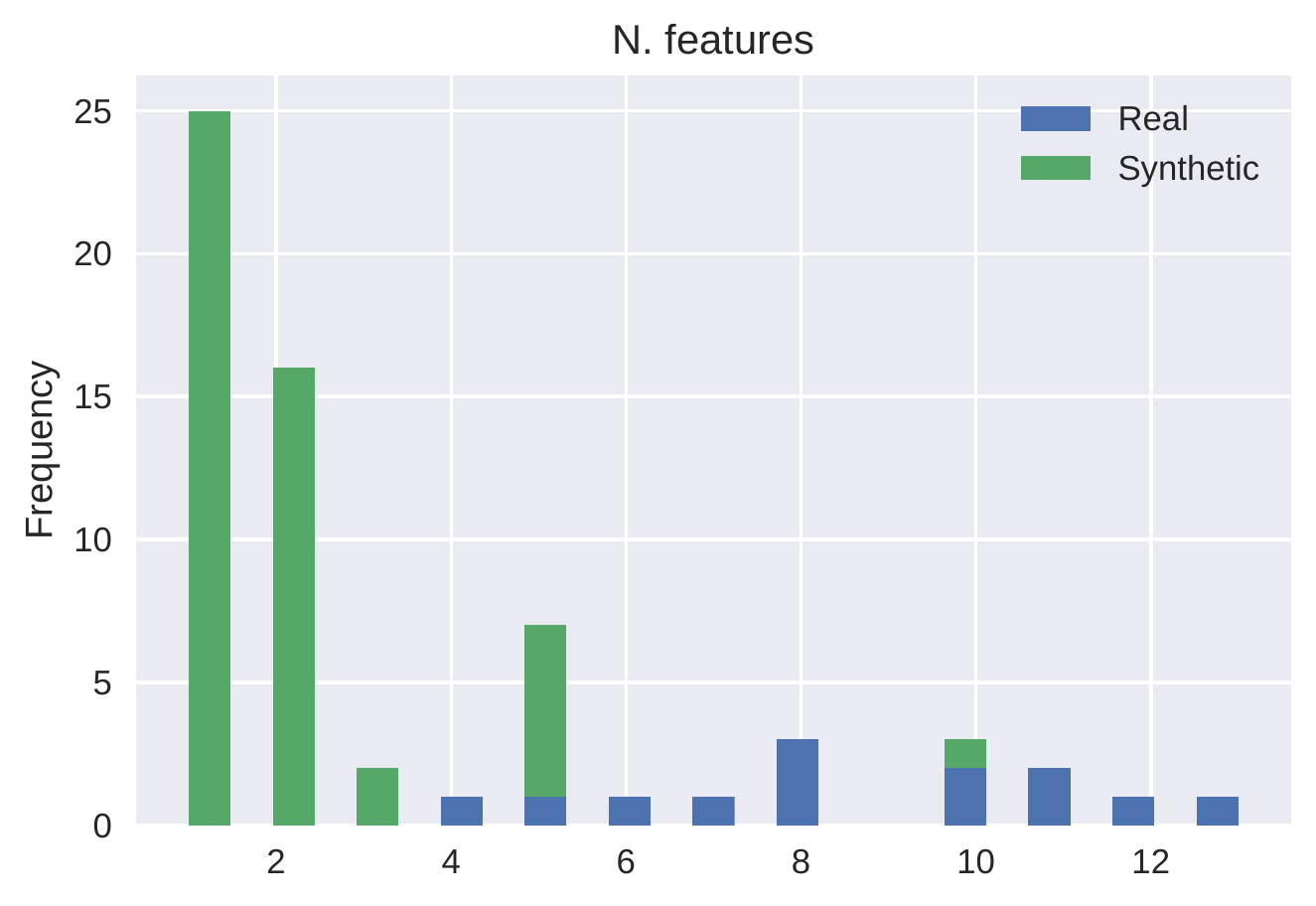}
		\caption{}
        \label{fig:hist-n-features}
    \end{subfigure}
    \hfill
    \begin{subfigure}[ht]{0.33\textwidth}
    	\centering
		\includegraphics[width=\linewidth]{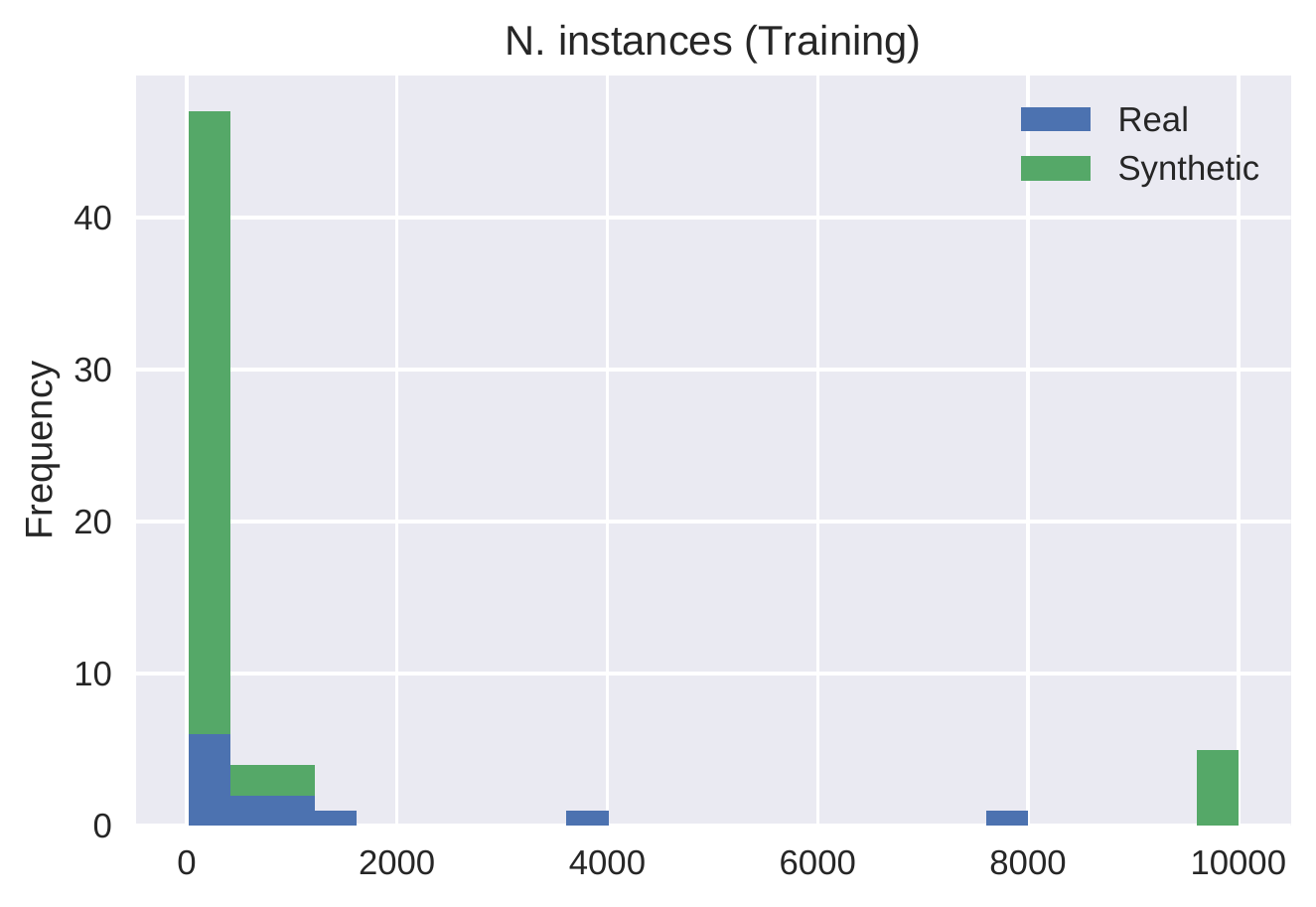}
		\caption{}		
		\label{fig:hist-n-instances-train}
	\end{subfigure}
    \hfill
    \begin{subfigure}[ht]{0.33\textwidth}
    	\centering
		\includegraphics[width=\linewidth]{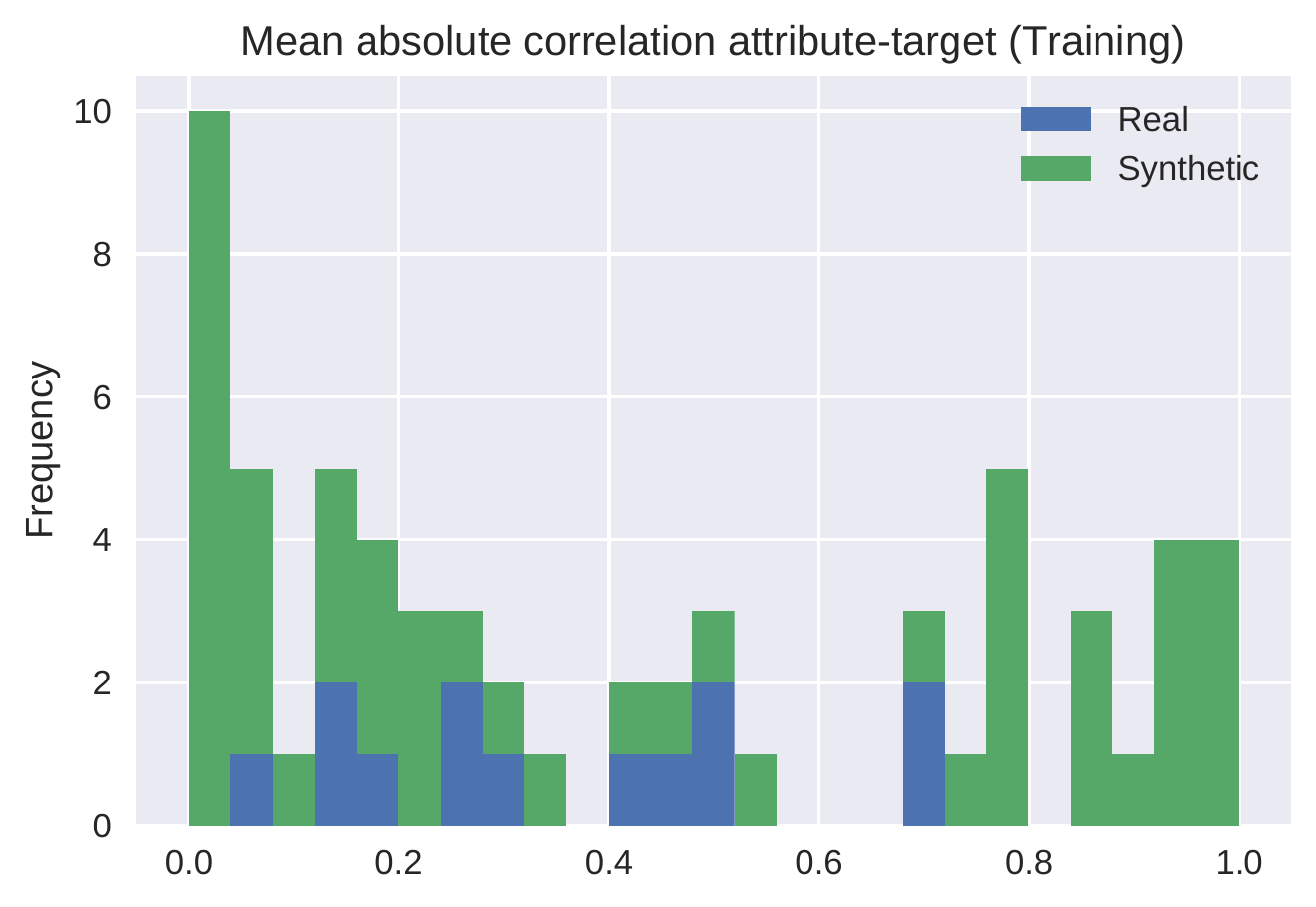}
		\caption{}		
		\label{fig:hist-corr-train}
	\end{subfigure}
    \hfill
    \begin{subfigure}[ht]{0.33\textwidth}
    	\centering
		\includegraphics[width=\linewidth]{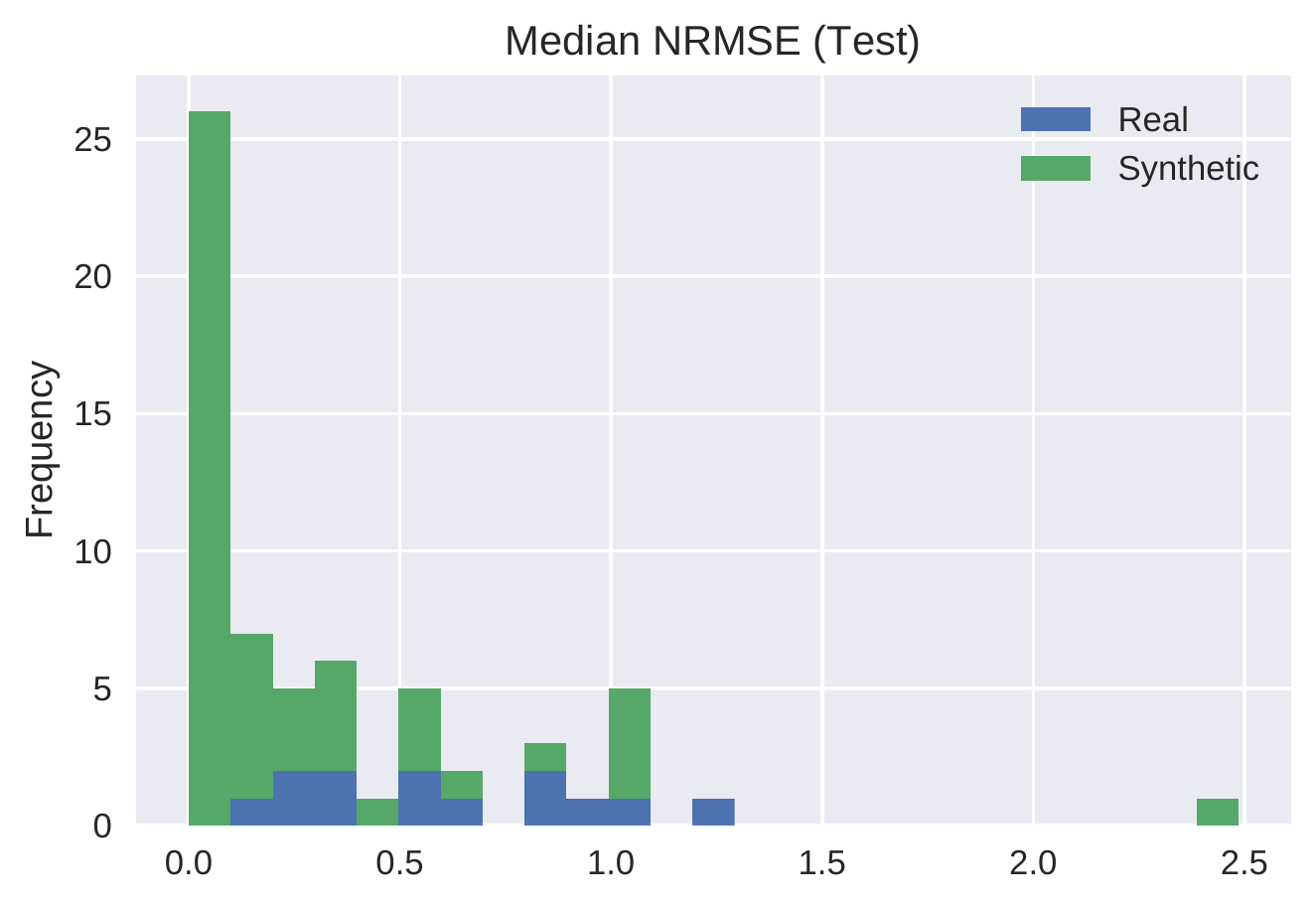}
		\caption{}		
		\label{fig:hist-nrmse-test}
	\end{subfigure}
    \hfill
    \begin{subfigure}[ht]{0.33\textwidth}
    	\centering
		\includegraphics[width=\linewidth]{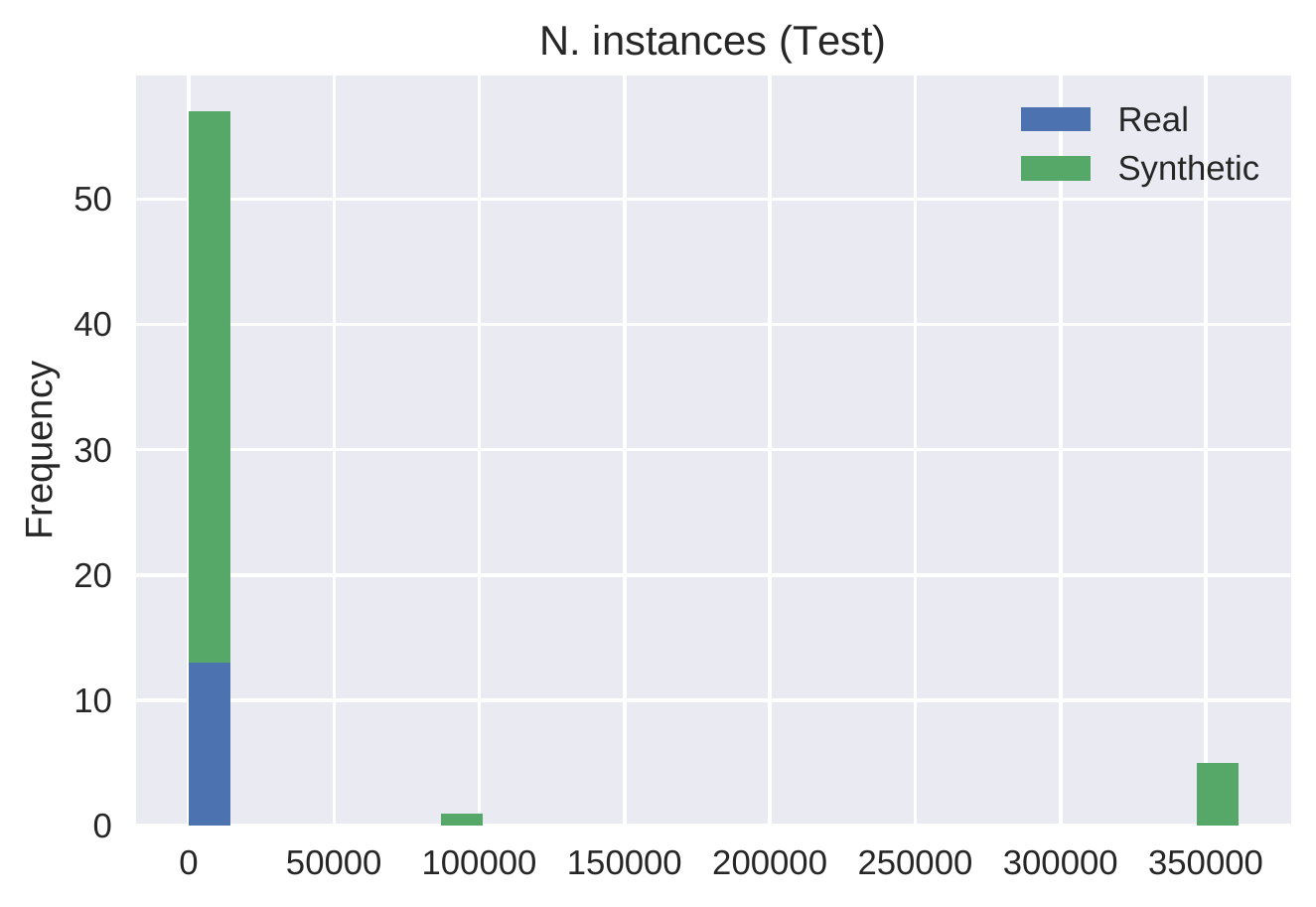}
		\caption{}		
		\label{fig:hist-n-instances-test}
	\end{subfigure}
    \hfill
    \begin{subfigure}[ht]{0.33\textwidth}
    	\centering
		\includegraphics[width=\linewidth]{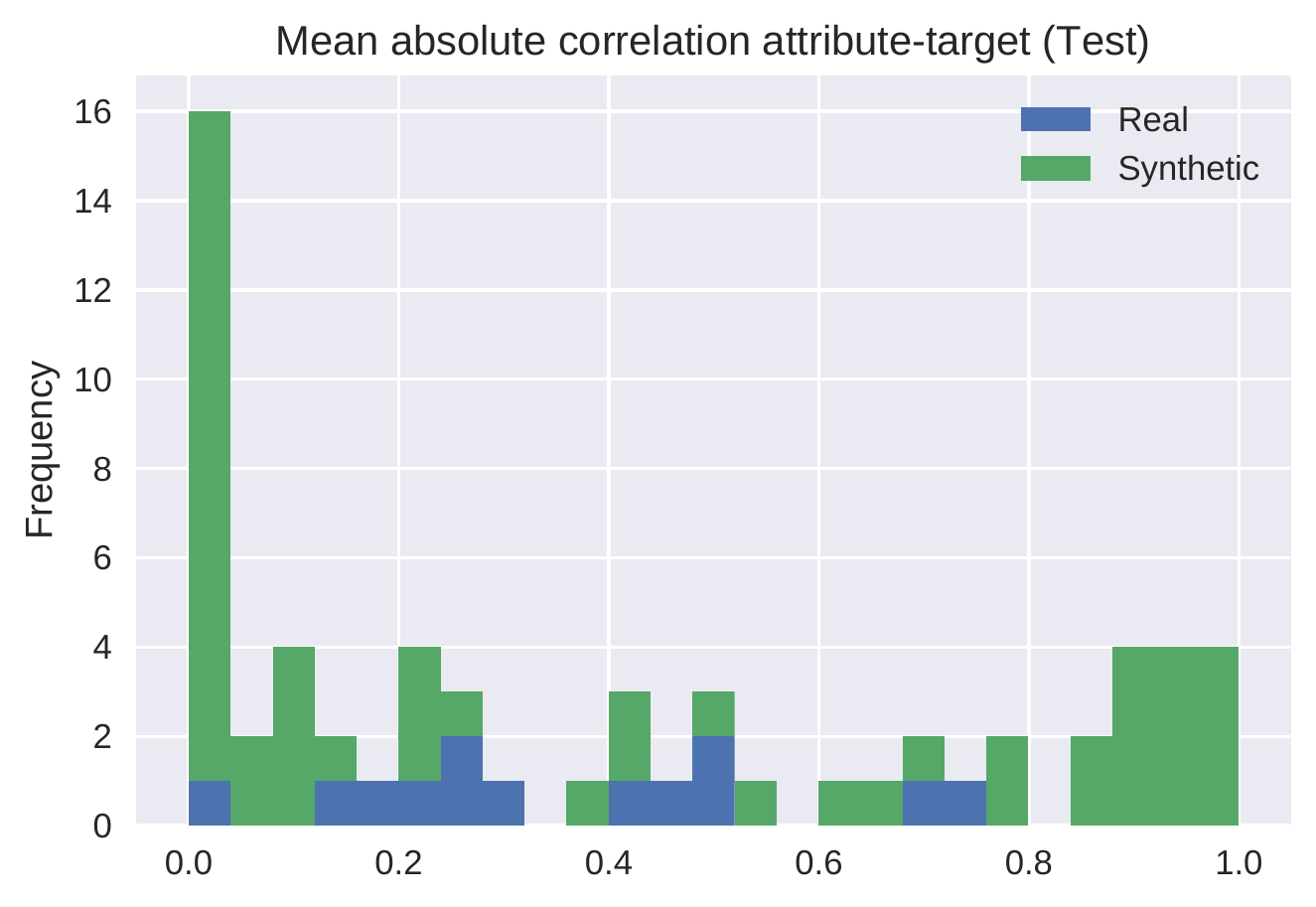}
		\caption{}		
		\label{fig:hist-corr-test}
	\end{subfigure}
    \hfill
	\begin{subfigure}[ht]{0.33\textwidth}
    	\centering
		\includegraphics[width=\linewidth]{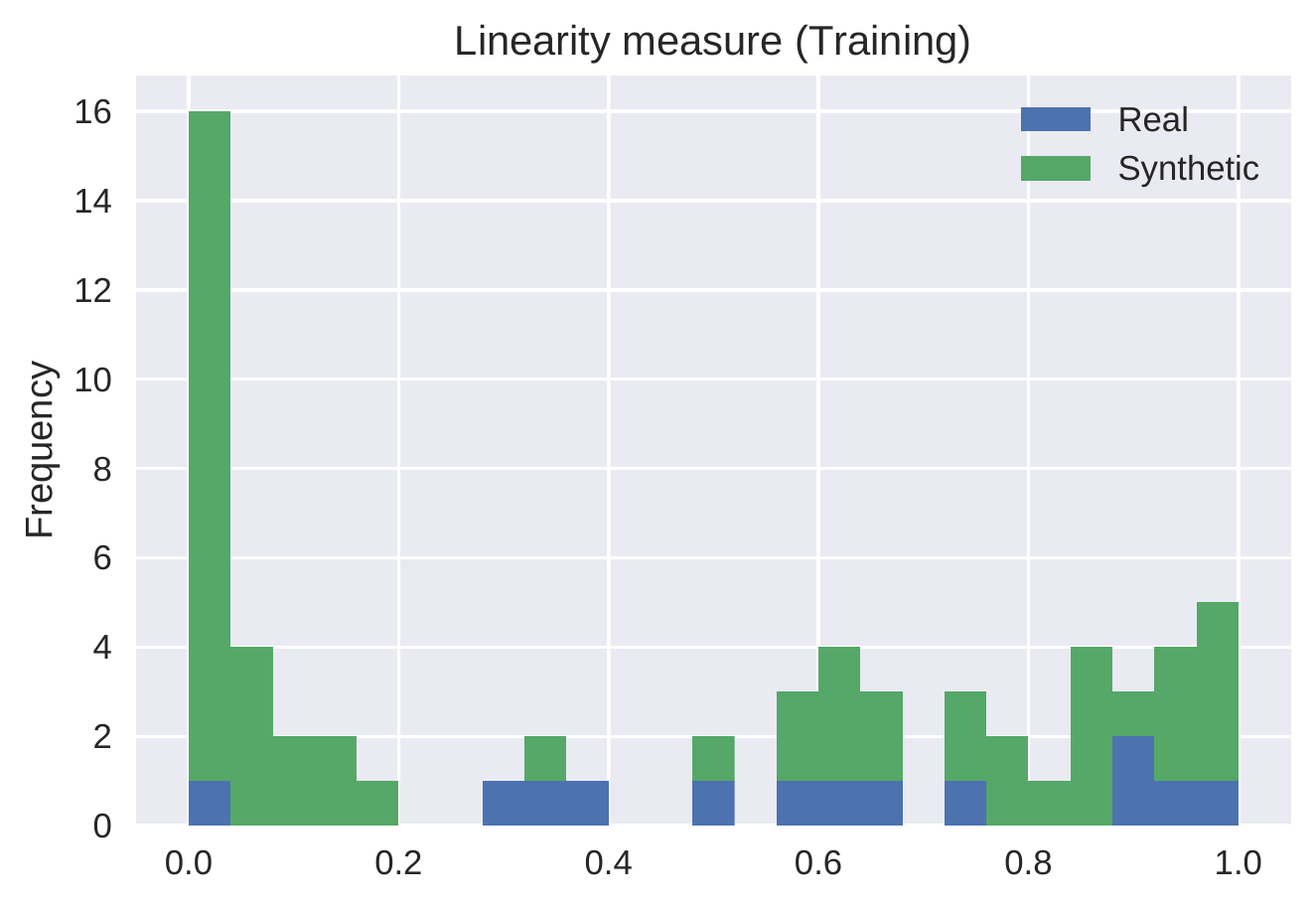}
		\caption{}
		\label{fig:hist-lin-train}
	\end{subfigure}
    \hfill
	\begin{subfigure}[ht]{0.33\textwidth}
    	\centering
		\includegraphics[width=\linewidth]{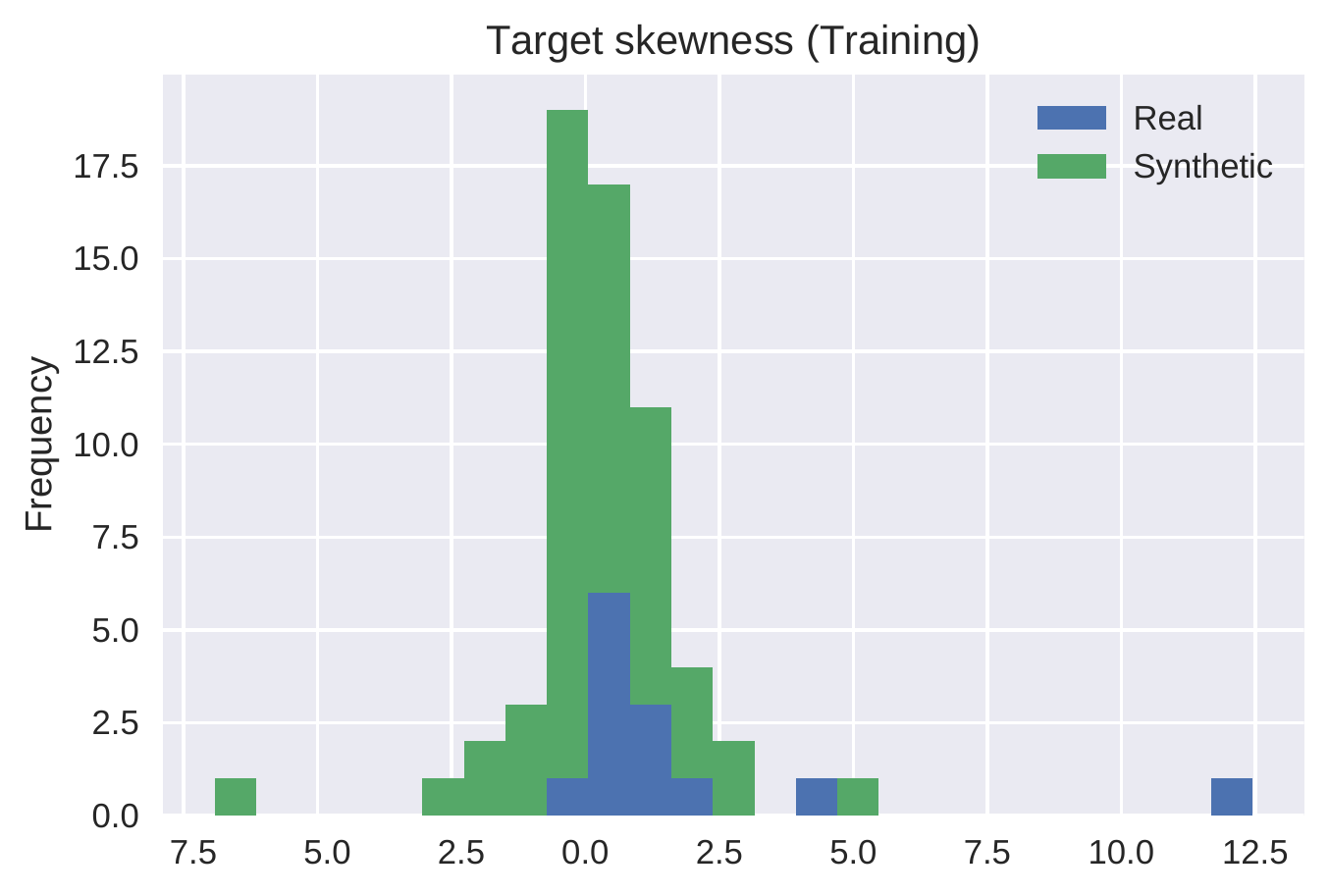}
		\caption{}		
		\label{fig:hist-skew-train}
	\end{subfigure}
    \hfill
    \begin{subfigure}[ht]{0.33\textwidth}
    	\centering
		\includegraphics[width=\linewidth]{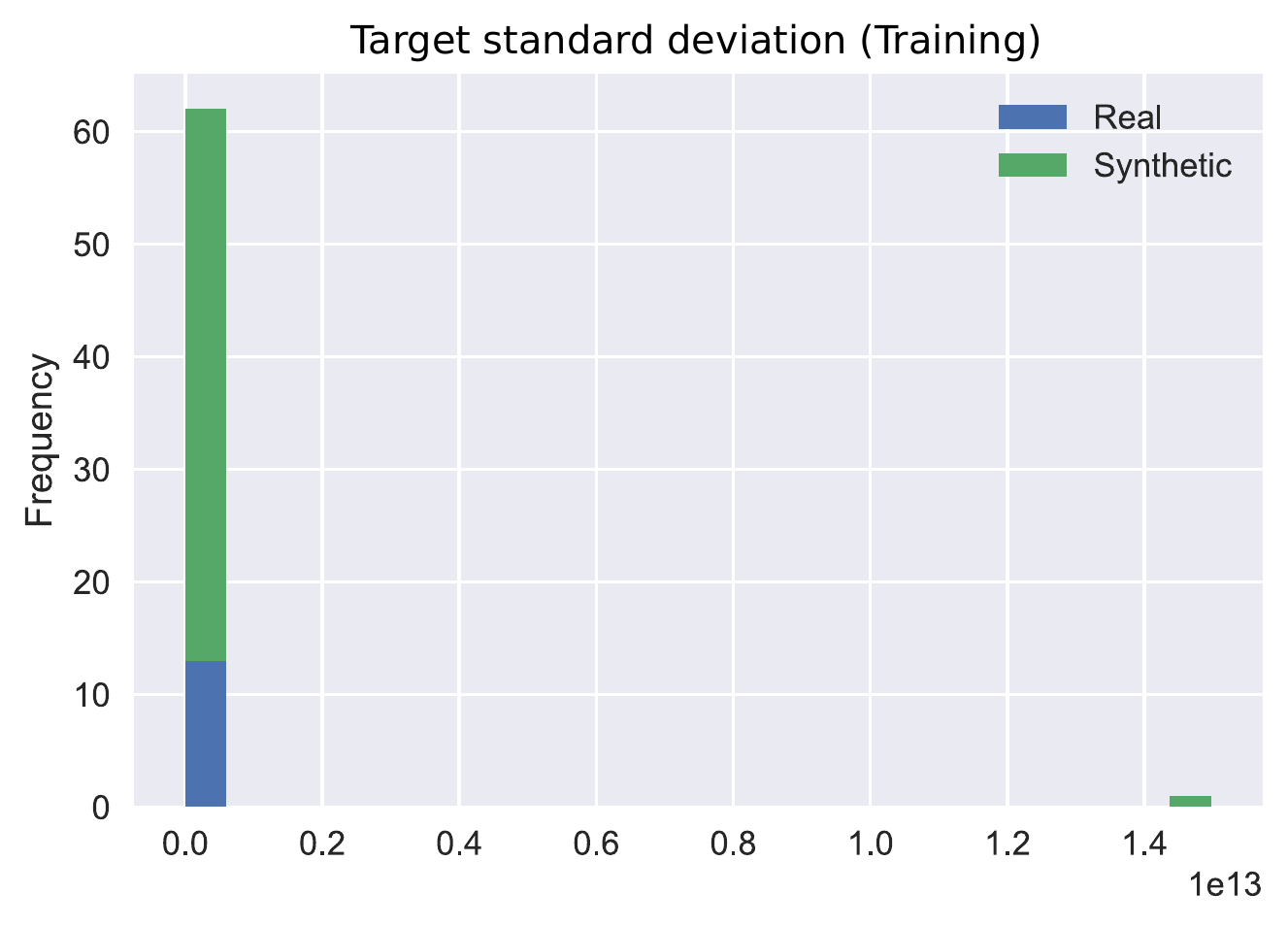}
		\caption{}		
		\label{fig:hist-target-std-train}
	\end{subfigure}
    \hfill
	\begin{subfigure}[ht]{0.33\textwidth}
    	\centering
		\includegraphics[width=\linewidth]{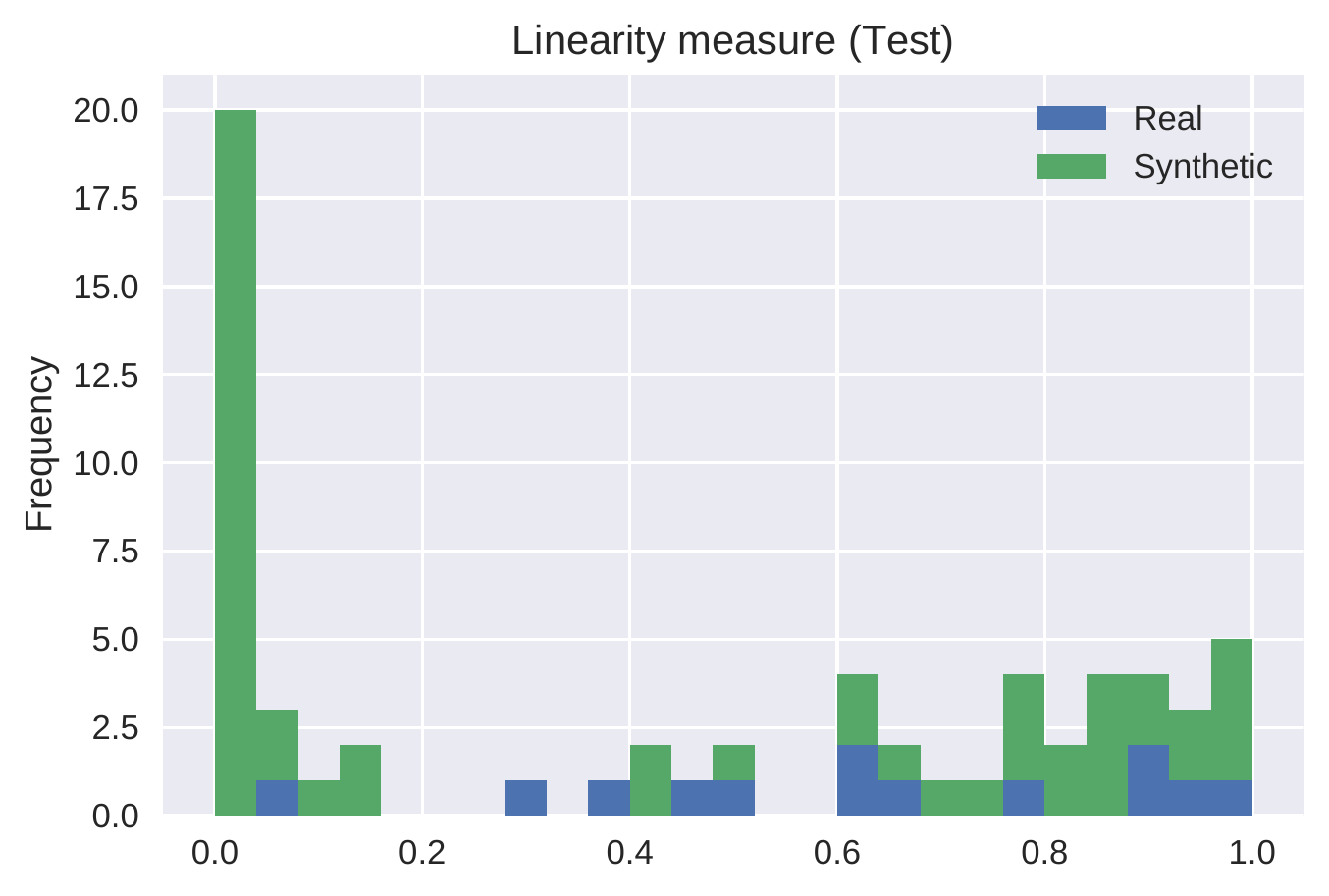}
		\caption{}
		\label{fig:hist-lin-test}
	\end{subfigure}
    \hfill
	\begin{subfigure}[ht]{0.33\textwidth}
    	\centering
		\includegraphics[width=\linewidth]{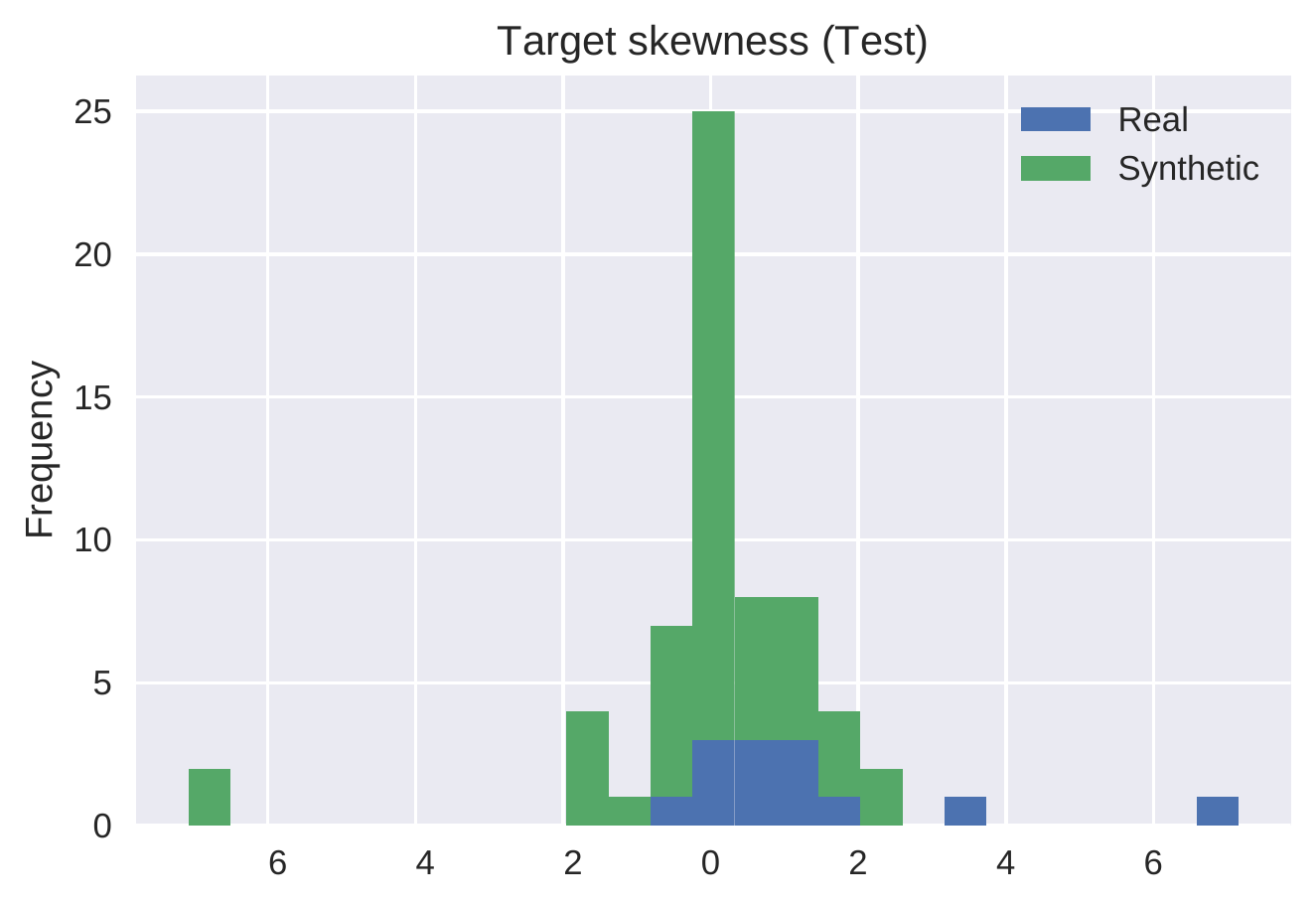}
		\caption{}		
		\label{fig:hist-skew-test}
	\end{subfigure}
    \hfill
    \begin{subfigure}[ht]{0.33\textwidth}
    	\centering
		\includegraphics[width=\linewidth]{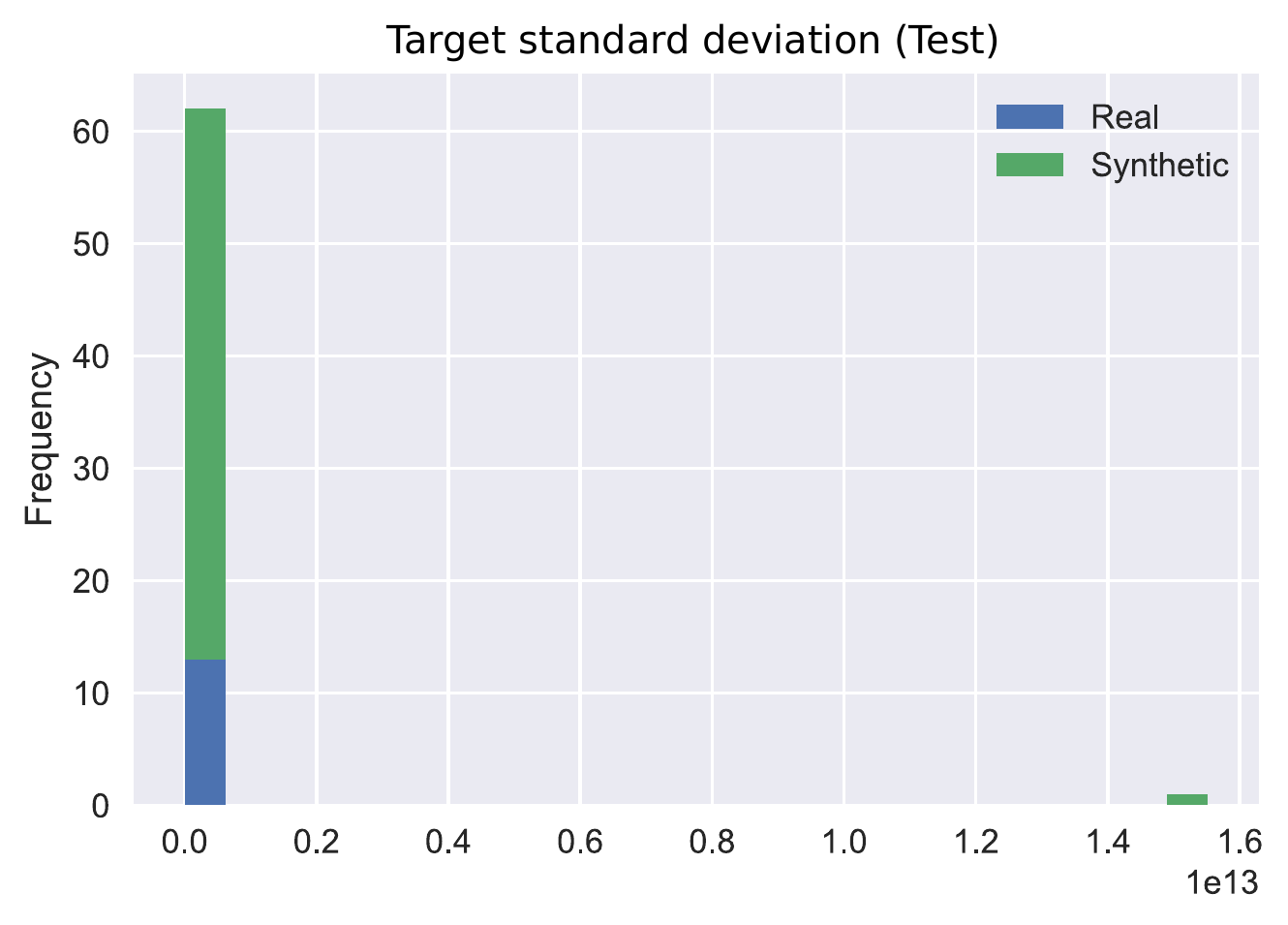}
		\caption{}		
		\label{fig:hist-target-std-test}
	\end{subfigure}
	\caption{Distribution of the values of the datasets meta-features across the 63 datasets considered.}
	\label{fig:hist}
\end{figure*}

Intuitively, a good set of benchmarks would include problems with a great variety of characteristics. Our methodology characterizes datasets according to a set of 11 meta-features, which we believe are relevant to show the relationship between the problem and the performance of GP, but they are far from being a complete set. Here we analyze the distribution of the values of these meta-features across the 63 datasets considered, and the results are illustrated in 
Figure~\ref{fig:hist}.
A general observation across all graphs is that there is a concentration of datasets in one point of this ``space'' that illustrates the benchmarks. For example, looking at the number of features (Figure~\ref{fig:hist-n-features}), we observed that the great majority of the datasets has one or two features, with the maximum number being 13. The same happens for the number of instances in the training and test sets (Figures~\ref{fig:hist-n-instances-train} and \ref{fig:hist-n-instances-test}), with the majority of datasets having less than 2,000 instances. Of course, that time restrictions need to be considered when creating benchmarks, but a few datasets with a greater number of examples can be beneficial to the analysis of the proposed method.

The mean absolute correlation of the predictive features and the target output (showed in Figures~\ref{fig:hist-corr-train} and \ref{fig:hist-corr-test}) shows that, individually, the features present, on average, very low correlation with the output. However, there is still 12 datasets with average correlation greater than 0.8 in the training set---\textit{Keijzer-6},  \textit{Keijzer-7}, \textit{Keijzer-8}, \textit{Keijzer-9}, \textit{Nguyen-1}, \textit{Nguyen-2}, \textit{Nguyen-3}, \textit{Nguyen-6}, \textit{Nguyen-7}, \textit{Nguyen-8}, \textit{R1} and \textit{R2}---and 13 datasets in the test set--- \textit{Keijzer-6}, \textit{Keijzer-7}, \textit{Keijzer-8}, \textit{Keijzer-9}, \textit{Nguyen-1}, \textit{Nguyen-2}, \textit{Nguyen-3}, \textit{Nguyen-4}, \textit{Nguyen-6}, \textit{Nguyen-7}, \textit{Nguyen-8}, \textit{Nonic}, \textit{R1} and \textit{R2}. As expected, also observe in these figures that the real datasets present lower average correlation than the synthetic ones.

Regarding our linearity measure represented by $R^2$ (Figures~\ref{fig:hist-lin-train} and \ref{fig:hist-lin-test}), observe that most datasets (46\% for the training and 49\% for the test set) have a $R^2$ lower than 0.5, which may be an indication that most of them are not described by linear relationships among the predictive features and the output. Finally, looking at the target skewness (Figures~\ref{fig:hist-skew-train} and \ref{fig:hist-skew-test}), notice that most of the datasets are concentrated from -1 to 1, with a peak around 0. A skewness value of 0 (meaning the data are perfectly symmetrical) is quite unlikely for real-world data, although we have seven real-world datasets---for both training and test set---with skewness in the interval [-0.5,0.5]. As a rule of thumb, datasets with skewness greater(smaller) than 1(-1) are considered highly skewed. We have 23(21) highly and 11(9) moderate skewed (with values between -1  and -0.5 and 0.5 and 1) training(test) datasets.



\begin{table}[!t]
  \small
  \rowcolors{2}{white}{gray!25}
  \caption{Meta-feature importance according to the Random Forest Regressor. The higher the value the more important is the feature.}
  \begin{center}
  \begin{tabular}{lc}
  	\toprule
    \textbf{Meta-Feature} & \textbf{Feat. import.} \\
    \midrule
      Target skewness (Training) & 0.290170 \\
      N. instances (Training) & 0.184041 \\
      N. features & 0.122526 \\
      Linearity measure (Training) & 0.075096 \\
      Target skewness (Test) & 0.056585 \\
      Target std (Training) & 0.054541 \\
      Linearity measure (Test) & 0.052174 \\
      Target std (Test) & 0.051961 \\
      Mean absolute correlation attribute-target (Test) & 0.044588 \\
      Mean absolute correlation attribute-target (Training) & 0.037164 \\
      N. instances (Test) & 0.031154 \\
	\midrule[\heavyrulewidth] 
\end{tabular}
\end{center}
\label{tab:feature-importance}
\end{table}

\subsection{Meta-features relevance}

In order to analyse the importance of each meta-feature on the prediction of GP performance---measured by the median NRMSE on the test set---we fitted the Random Forest Regressor from Scikit-learn \cite{pedregosa2011scikit} to the meta-dataset---composed of the meta-features of all datasets as input and the respective median NRMSE as output. Table \ref{tab:feature-importance} presents the feature importance inferred by the Random Forest Regressor from Scikit-learn \cite{pedregosa2011scikit}. The number of trees in the forest was set to 120, according to a parameter tuning using the randomized search from Scikit-learn---all the other parameters were set to the default value. 
Notice that the target skewness and the number of instances in the training set were considered the most relevant meta-features, followed by the number of features.

\begin{figure}[t]
	\centering
	\begin{subfigure}[ht]{0.48\textwidth}
		\includegraphics[width=\linewidth]{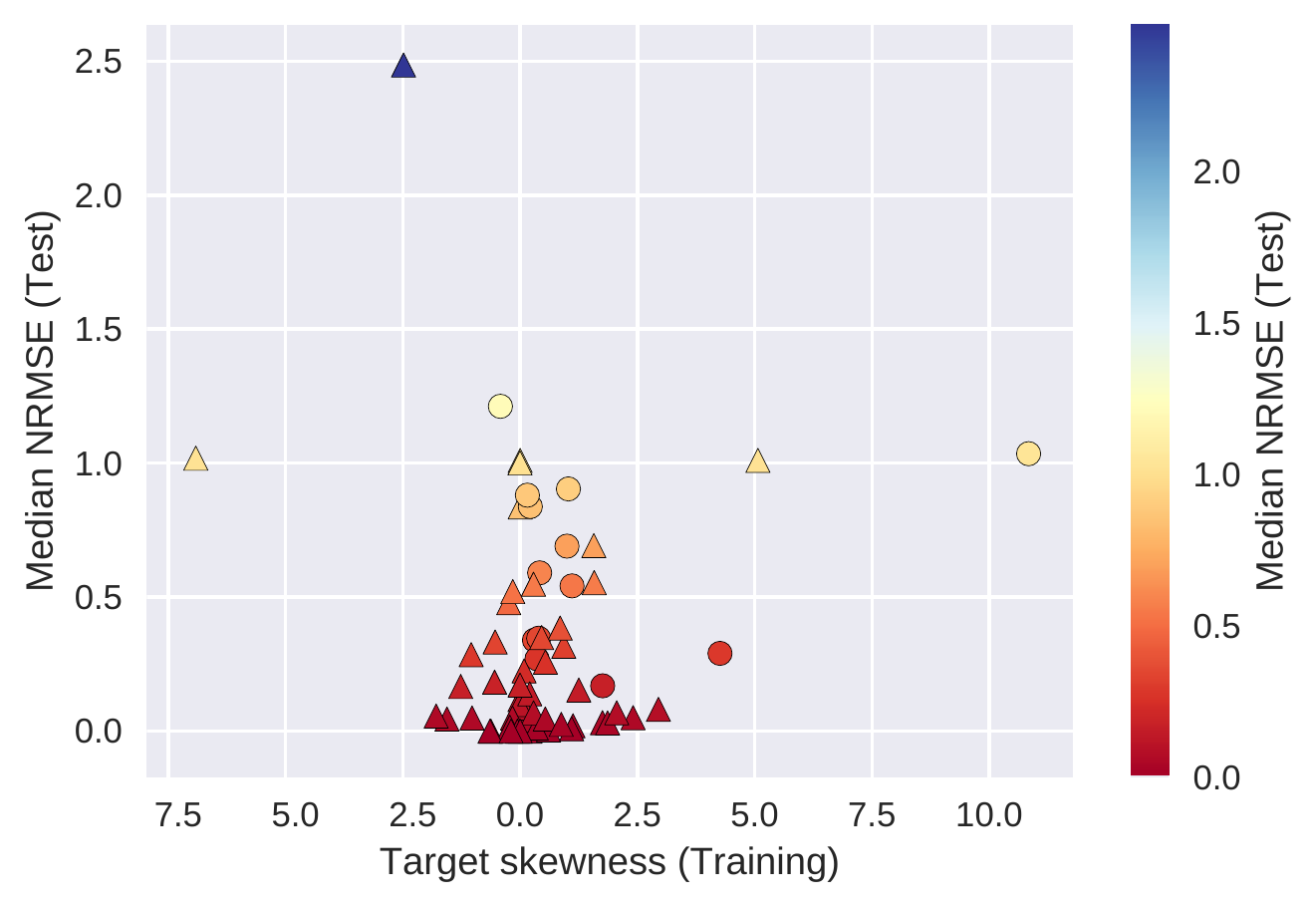}
		\caption{}
		\label{fig:skew-vs-nrmse}
	\end{subfigure}
	\hfill
	\begin{subfigure}[ht]{0.48\textwidth}
		\includegraphics[width=\linewidth]{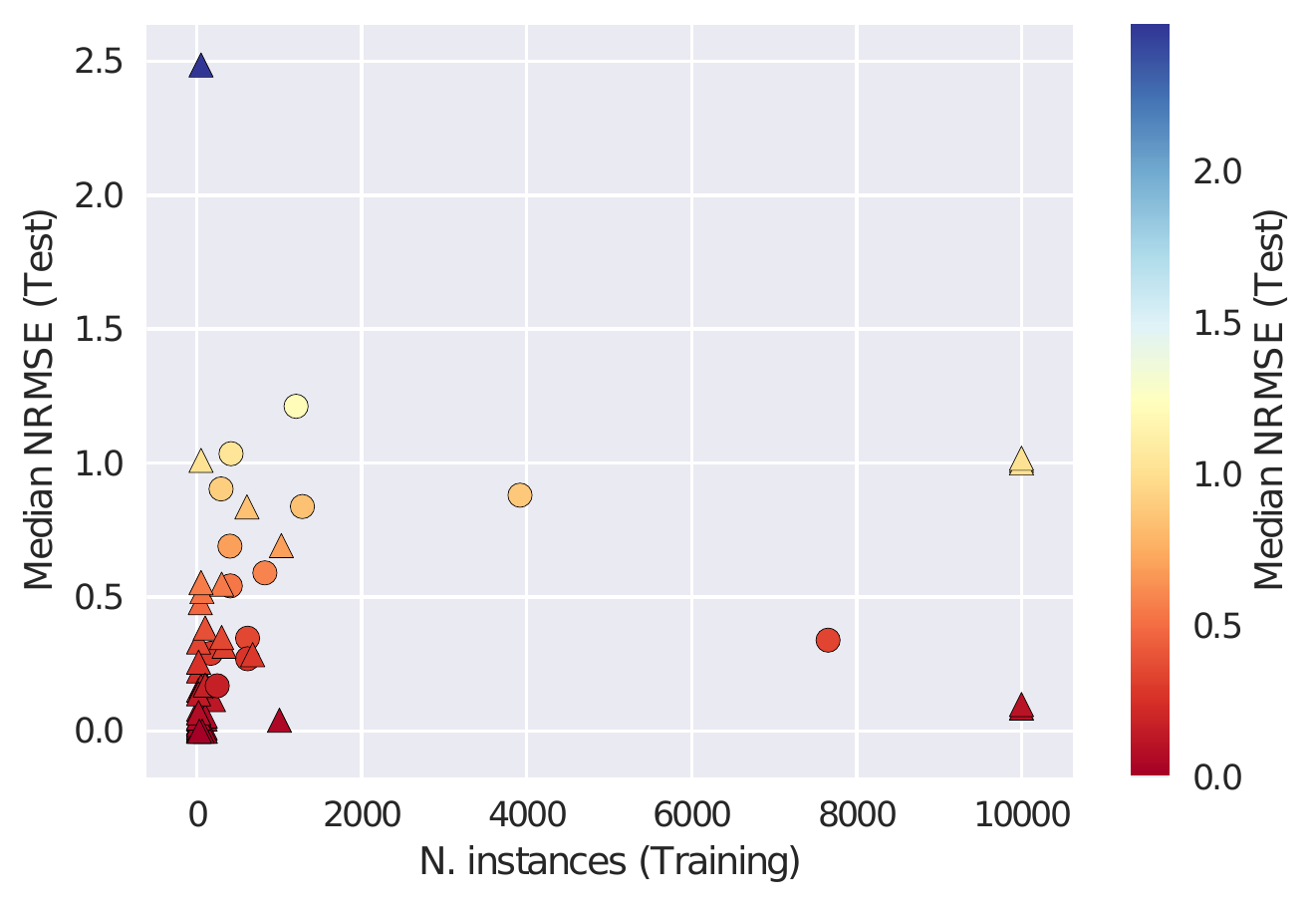}
		\caption{}
		\label{fig:n-feat-vs-nrmse}
	\end{subfigure}
	\caption{Relation between the (a) the target skewness in the training set and (b) the number of instances in the training set compared to the median test NRMSE. Circles represent real datasets, while triangles represent synthetic ones.}
	\label{fig:skew-n-inst-vs-nrmse}
\end{figure}

Figure \ref{fig:skew-n-inst-vs-nrmse} presents the values of NRMSE in the test set according to the two most important features according to Table \ref{tab:feature-importance}. Circles correspond to real-world datasets and triangles to synthetic ones. The colour of each element and its position on the vertical axis represent the median test NRMSE obtained as output by the regression models. 
Note that the concentration of datasets with skewness close to 0 leads to very small errors in these datasets, which are mostly synthetic ones. For the number of instances, we observe that smaller datasets tend to have smaller error, especially for synthetic datasets---with the exception of two datasets, \textit{Korns-1} and \textit{Korns-4}, which present small NRMSE and have 10,000 training instances.

\begin{figure}[t]
	\centering
	\includegraphics[width=\linewidth]{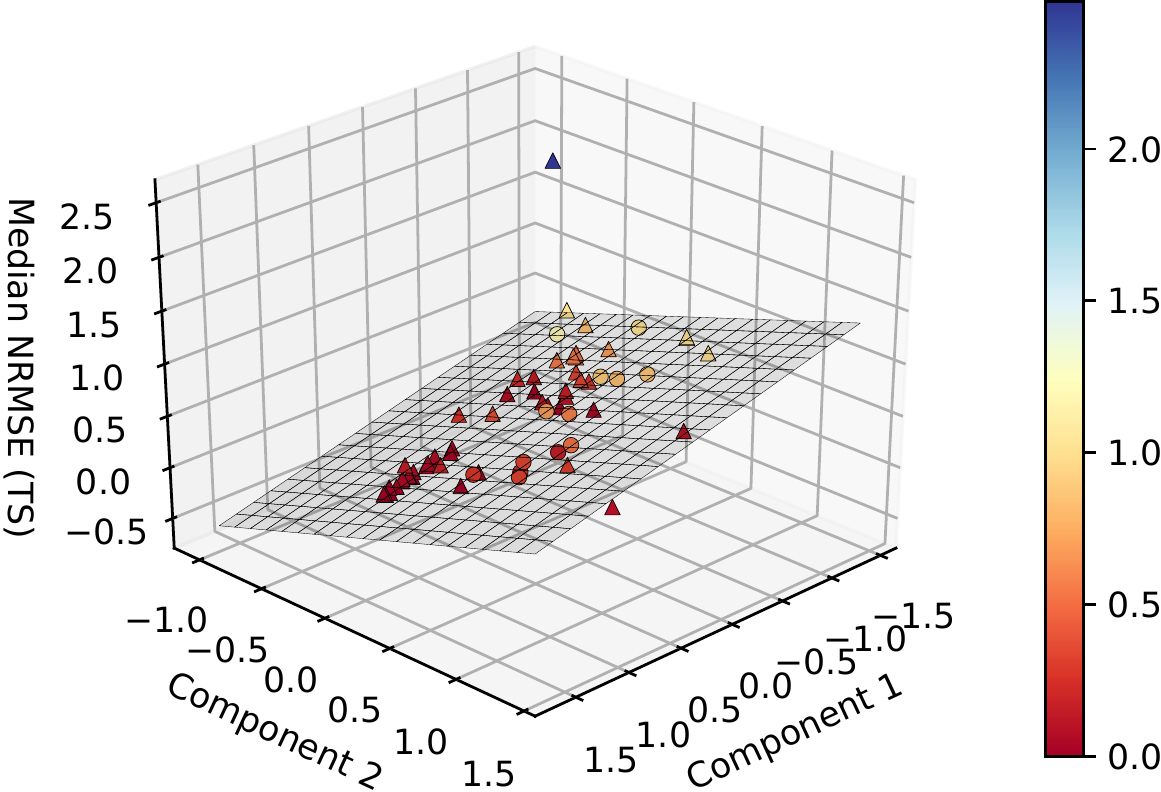}
	\caption[Relation between the median test NRMSE and meta-features mapped to a two-dimensional space using the PCA method.]{Relation between the median test NRMSE and meta-features mapped to a two-dimensional space using the PCA method. Circles represent real datasets, while triangles represent synthetic ones.}
	\label{fig:pca-lr}
\end{figure}

In order to be able to visualize where the datasets fall in the dataset space generated by the meta-features, we plotted in Figure~\ref{fig:pca-lr} the two principal components of PCA---induce from the meta-features normalized to $[0,1]$---together with the test NRMSE returned by GP.
Again, circles correspond to real-world datasets and triangles to synthetic ones, and the colour of each element and its position on the vertical axis represent the median test NRMSE obtained as output by the GP. The figure also shows the least-squares plane generated using a linear regression method. Observe that all datasets, with a few exceptions, are really concentrated in a portion of the space generated by these two components. In an ideal scenario of benchmarks, there should be a better distribution of these datasets in this space.
The plane generated using these two components as a representation for the meta-features obtained a coefficient of determination of 0.332 and a RMSE of 0.356. 

If the same linear regression is applied to the full set of meta-features, the coefficient of determination increases to 0.491 and the RMSE decreases to 0.311. However, the Random Forest Regressor model, that generated the feature importance previously reported, had a coefficient of determination of 0.894 and a RMSE of 0.142, giving evidence that the relations among the predictive features and the target output are mostly non-linear. 

	\section{Conclusions and Future Work}
\label{sec:conclusions}

The main contribution of this paper was to move the GP community towards the proposal of a framework to measure the quality of the benchmarks quantitatively.
We started by performing an analysis of the datasets commonly used as benchmarks for symbolic regression in the GP scenario by using a meta-learning approach. This approach extracted a set of meta-features from the benchmark datasets, and analysed how they correlate to the output error of a canonical GP. We extracted 11 meta-features from 63 datasets, analysed how the later were distributed according to the former and also reduced these 11 meta-features to two using a dimensionality reduction method. In this way, we were able to visualize the meta-feature space and observe that most examples were concentrated in the same region of this induced space. This suggests that the datasets are still very similar, and more variety is desired in a set of benchmarks.

As future work, we want to first explore a wider set of meta-features, and add new datasets to our meta-dataset. We also intend to analyse different regression methods to build an improved model for the NRMSE prediction from meta-features.

	\begin{acks}

      This work was partially supported by the following Brazilian Research Support Agencies: CNPq, FAPEMIG, CAPES. It has also been partially funded by the EUBra-BIGSEA project by the European Commission under the Cooperation Programme (MCTI/RNP 3rd Coordinated Call), Horizon 2020 grant agreement 690116.
		
	\end{acks}
	
	\bibliographystyle{ACM-Reference-Format}
	\bibliography{references}
    
    \begin{table*}[ht]
    \scriptsize
    \rowcolors{2}{white}{gray!25}
    \caption{Synthetic datasets used in GECCO papers from 2013-2017.}
    \begin{center}
        \begin{tabular}{lclllr}
            \toprule
            \textbf{Dataset} & \textbf{Variables} & \textbf{Objective Function} & \textbf{Training Set} & \textbf{Testing Set} & \textbf{Source}\\
            \cmidrule(r{0em}){1-6}

\rowcolor{gray!25}            Burks (*) & 1 & $4*x_1^4+3*x_1^3+2*x_1^2+x_1$ & $U[-1,1,20]$ & None & \cite{burks2015efficient,szubert2016reducing}\\

\rowcolor{white}            Keijzer-1 & 1 & $0.3*x^1*sin(2*\pi*x_1)$ & $E[-1, 1, 0.1]$ & $E[-1, 1, 0.001]$ & \cite{krawiec2013approximating,krawiec2014behavioral,demelo2014kaizen,miranda2017how,liskowski2017discovery}\\
\rowcolor{gray!25}          Keijzer-2 & 1 & $0.3*x^1*sin(2*\pi*x_1)$ & $E[-2, 2, 0.1]$ & $E[-2, 2, 0.001]$ & \cite{demelo2014kaizen,miranda2017how}\\
\rowcolor{white}            Keijzer-3 & 1 & $0.3*x^1*sin(2*\pi*x_1)$ & $E[-3, 3, 0.1]$ & $E[-3, 3, 0.001]$ & \cite{demelo2014kaizen,miranda2017how}\\
\rowcolor{gray!25}          Keijzer-4 & 1 & $x_1^3*e^{-x_1}*cos(x_1)*sin(x_1)*(sin^2(x_1)*cos(x_1)-1)$ & $E[0, 10, 0.1]$ & $E[0.05, 10.05, 0.1]$ & \cite{krawiec2013approximating,krawiec2014behavioral,demelo2014kaizen,szubert2016reducing,chen2016improving,sotto2017probabilistic,miranda2017how,liskowski2017discovery}\\

\rowcolor{white} &  &  & $x_1, x_2: U[-1, 1, 1000]$ & $x_1, x_2: U[-1, 1, 10000]$ & \\
 \multirow{-2}{*}{Keijzer-5} & \multirow{-2}{*}{3} & \multirow{-2}{*}{$\frac{30*x_1*x_3}{(x_1-10)*x_2^2}$} & $x_3: U[1, 2, 1000]$ & $x_3: U[1, 2, 10000]$ & \multirow{-2}{*}{\cite{krawiec2014behavioral,demelo2014kaizen,sotto2017probabilistic,oliveira2016dispersion}} \\

            Keijzer-6 & 1 & $\sum_{i=1}^{x_1}i$ & $E[1, 50, 1]$ & $E[1, 120, 1]$ & \cite{demelo2014kaizen,lacava2015genetic,nicolau2016managing,oliveira2016dispersion,miranda2017how,medvet2017evolvability}\\
            Keijzer-7 & 1 & $ln x_1$ & $E[1, 100, 1]$ & $E[1, 100, 0.1]$ & \cite{demelo2014kaizen,oliveira2016dispersion,miranda2017how}\\
            Keijzer-8 & 1 & $\sqrt{x_1}$ & $E[0, 100, 1]$ & $E[0, 100, 0.1]$ & \cite{demelo2014kaizen,miranda2017how,liskowski2017discovery}\\
            Keijzer-9 & 1 & $arcsinh(x_1) = ln(x_1+\sqrt{x_1^2+1})$ & $E[0, 100, 1]$ & $E[0, 100, 0.1]$ & \cite{demelo2014kaizen,miranda2017how}\\
            Keijzer-10 & 2 & $x_1^{x_2}$ & $U[0, 1, 100]$ & $E[0, 1, 0.01]$ & \cite{wieloch2013running,demelo2014kaizen,thuong2017combining}\\
            Keijzer-11 & 2 & $x_1*x_2+sin((x_1-1)*(x_2-1))$ & $U[-3, 3, 20]$ & $E[-3, 3, 0.01]$ & \cite{krawiec2014behavioral,demelo2014kaizen,chen2016improving,thuong2017combining} \\
            Keijzer-12 (*)& 2 & $x_1^4-x_1^3+(\frac{x_2^2}{2})-x_2$ & $U[-3, 3, 20]$ & $E[-3, 3, 0.01]$ & \cite{wieloch2013running,krawiec2014behavioral,demelo2014kaizen,chen2016improving,thuong2017combining}\\
            Keijzer-13 & 2 & $6*sin(x_1)*cos(x_2)$ & $U[-3, 3, 20]$ & $E[-3, 3, 0.01]$ & \cite{krawiec2014behavioral,demelo2014kaizen,thuong2017combining}\\
            Keijzer-14 & 2 & $\frac{8}{2+x_1^2+x_2^2}$ & $U[-3, 3, 20]$ & $E[-3, 3, 0.01]$ & \cite{krawiec2014behavioral,demelo2014kaizen,chen2016improving,liskowski2017discovery,thuong2017combining}\\
            Keijzer-15 & 2 & $\frac{x_1^3}{5}+\frac{x_2^3}{2}-x_2-x_1$ & $U[-3, 3, 20]$ & $E[-3, 3, 0.01]$ & \cite{krawiec2014behavioral,demelo2014kaizen,chen2016improving,liskowski2017discovery,thuong2017combining}\\

            Korns-1 & 1 & $1.57+24.3*x_4$ & $U[-50,50,10000]$ & $U[-50,50,10000]$ & \cite{worm2013prioritized}\\
            Korns-2\textsuperscript{1} & 3 & $0.23+14.2*\frac{x_4+x_2}{3*x_5}$ & $U[-50,50,10000]$ & $U[-50,50,10000]$ & \cite{worm2013prioritized}\\
            Korns-3\textsuperscript{1} & 4 & $-5.41+4.9*\frac{x_4-x_1+\frac{x_2}{x_5}}{3*x_5}$ & $U[-50,50,10000]$ & $U[-50,50,10000]$ & \cite{worm2013prioritized,sotto2017probabilistic}\\
            Korns-4 & 1 & $-2.3+0.13*sin(x_3)$ & $U[-50,50,10000]$ & $U[-50,50,10000]$ & \cite{worm2013prioritized}\\
            Korns-5\textsuperscript{1} & 1 & $3+2.13*ln(x_5)$ & $U[-50,50,10000]$ & $U[-50,50,10000]$ &  \cite{worm2013prioritized,sotto2017probabilistic}\\
            Korns-6\textsuperscript{1} & 1 & $1.3+0.13*\sqrt{x_1}$ & $U[-50,50,10000]$ & $U[-50,50,10000]$ & \cite{worm2013prioritized}\\
            Korns-7 & 1 & $213.80940889*(1-e^{-0.54723748542*x_1})$ & $U[-50,50,10000]$ & $U[-50,50,10000]$ & \cite{worm2013prioritized}\\
            Korns-8\textsuperscript{1} & 3 & $6.87+11*\sqrt{7.23*x_1*x_4*x_5}$ & $U[-50,50,10000]$ & $U[-50,50,10000]$ & \cite{worm2013prioritized}\\
            Korns-9\textsuperscript{1} & 4 & $\frac{\sqrt{x_1}}{ln(x_2)}*\frac{e^{x_3}}{x_4^2}$ & $U[-50,50,10000]$ & $U[-50,50,10000]$ & \cite{worm2013prioritized}\\
            Korns-10\textsuperscript{1} & 4 & $0.81+24.3*\frac{2*x_2+3*x_3^2}{4*x_4^3+5*x_5^4}$ & $U[-50,50,10000]$ & $U[-50,50,10000]$ & \cite{worm2013prioritized}\\
            Korns-11 & 1 & $6.87+11*cos(7.23*x_1^3)$ & $U[-50,50,10000]$ & $U[-50,50,10000]$ & \cite{worm2013prioritized}\\
            Korns-12 & 2 & $2-2.1*cos(9.8*x_1)*sin(1.3*x_5)$ & $U[-50,50,10000]$ & $U[-50,50,10000]$ & \cite{worm2013prioritized}\\

            Koza-2 (*)& 1 & $x_1^5-2*x_1^3+x_1$ & $U(-1, 1, 20)$ & None & \cite{meier2013accelerating}\\ 
            Koza-3 (*)& 1 & $x_1^6-2*x_1^4+x_1^2$ & $U(-1, 1, 20)$ & None & \cite{meier2013accelerating, harada2014asynchronously}\\ 

            Meier-3  & 2 & $\frac{x_1^2*x_2^2}{x_1+x_2}$ & $U[-1,1,50]$ & None & \cite{meier2013accelerating}\\
            Meier-4  & 2 & $\frac{x_1^5}{x_2^3}$ & $U[-1,1,50]$ & None & \cite{meier2013accelerating}\\

            Nguyen-1 (*)& 1 & $x_1^3+x_1^2+x_1$ & $U(-1, 1, 20)$ & None & \cite{worm2013prioritized,demelo2014kaizen,sotto2017probabilistic}\\
            Nguyen-2 (*)& 1 & $x_1^4+x_1^3+x_1^2+x_1$ & $U(-1, 1, 20)$ & None & \cite{worm2013prioritized,lopes2013gearnet,harada2014asynchronously,demelo2014kaizen,whigham2015examining,sotto2017probabilistic,medvet2017evolvability}\\
            Nguyen-3 & 1 & $x_1^5+x_1^4+x_1^3+x_1^2+x_1$ & $U(-1, 1, 20)$ & None & \cite{worm2013prioritized,wieloch2013running,krawiec2014behavioral,demelo2014kaizen,sotto2017probabilistic,liskowski2017discovery}\\
            Nguyen-4 & 1 & $x_1^6+x_1^5+x_1^4+x_1^3+x_1^2+x_1$ & $U(-1, 1, 20)$ & None & \cite{worm2013prioritized,wieloch2013running,krawiec2014behavioral,demelo2014kaizen,sotto2017probabilistic,liskowski2017discovery}\\
            Nguyen-5 & 1 & $sin(x_1^2)*cos(x_1)-1$ & $U(-1, 1, 20)$ & None & \cite{worm2013prioritized,wieloch2013running,harada2014asynchronously,krawiec2014behavioral,demelo2014kaizen,liskowski2017discovery}\\
            Nguyen-6 & 1 & $sin(x_1)+sin(x_1+x_1^2)$ & $U(-1, 1, 20)$ & None & \cite{worm2013prioritized,wieloch2013running,krawiec2014behavioral,demelo2014kaizen,sotto2017probabilistic,liskowski2017discovery}\\
            Nguyen-7 & 1 & $ln(x_1+1)+ln(x_1^2+1)$ & $U[0,2,20]$ & None & \cite{krawiec2013approximating},\cite{worm2013prioritized,wieloch2013running,harada2014asynchronously,krawiec2014behavioral,demelo2014kaizen,lacava2015genetic,liskowski2017discovery}\\ 
            Nguyen-8 & 1 & $\sqrt{x_1}$ & $U[0,4,20]$ & None & \cite{worm2013prioritized,wieloch2013running,demelo2014kaizen,liskowski2017discovery}\\ 
            Nguyen-9 & 2 & $sin(x_1)+sin(x_2^2)$ & $U(-1, 1, 100)$ & None & \cite{worm2013prioritized,wieloch2013running,krawiec2014behavioral,demelo2014kaizen,liskowski2017discovery}\\
            Nguyen-10 & 2 & $2*sin(x_1)*cos(x_2)$ & $U(-1, 1, 100)$ & None & \cite{worm2013prioritized,wieloch2013running,krawiec2014behavioral,demelo2014kaizen}\\

            Nonic & 1 & $\sum_{i=1}^9 x_1^i$ & $E[-1,1,20]$ & $U[-1,1,20]$ & \cite{krawiec2013approximating,szubert2016reducing}\\ 
            
            Pagie-1 & 2 & $\frac{1}{1+x_1^{-4}}+\frac{1}{1+x_2^{-4}}$ & $E[-5, 5, 0.4]$ & None & \cite{mcphee2015impact,lacava2015genetic,liskowski2017discovery}\\ 
        
            Poly-10 & 9 & $x_1*x_2+x_3*x_4+x_5*x_6+x_1*x_7*x_9+x_3*x_6*x_{10}$ & $U[0,1,330]$ & $U[0,1,170] $ &  \cite{medernach2016new}\\ 
            
            R1 & 1 & $\frac{(x_1+1)^3}{x_1^2-x_1+1}$ & $E[-1,1,20]$ & $U[-1,1,20]$ & \cite{krawiec2013approximating,szubert2016reducing,liskowski2017discovery}\\ 
            R2 & 1 & $\frac{x_1^5-3*x_1^3+1}{x_1^2+1}$ & $E[-1,1,20]$ & $U[-1,1,20]$ & \cite{krawiec2013approximating,szubert2016reducing,liskowski2017discovery}\\ 
            R3 & 1 & $\frac{x_1^6+x_1^5}{x_1^4+x_1^3+x_1^2+x_1+1}$ & $E[-1,1,20]$ & $U[-1,1,20]$ & \cite{liskowski2017discovery}\\ 

            Sine & 1 & $sin(x_1)+sin(x_1+x_1^2)$ & $E[0,6.2,0.1]$ & None & \cite{mcphee2015impact}\\ 
            
            Vladislavleva-1 & 2 & $\frac{e^{-(x_1-1)^2}}{1.2+(x_2-2.5)^2}$ & $U[0.3, 4, 100]$ & $E[-0.2, 4.2, 0.1]$ & \cite{oliveira2016dispersion,miranda2017how,liskowski2017discovery,thuong2017combining}\\ 
            Vladislavleva-2 & 1 & $e^{-x_1}*x_1^3*(cos x_1*sin x_1)*(cos x_1*sin^2 x_1-1)$ & $E[0.05, 10, 0.1]$ & $E[-0.5, 10.5, 0.05]$ & \cite{miranda2017how}\\ 

& & & $x_1: E[0.05, 10, 0.1]$ & $x_1: E[-0.5, 10.5, 0.05]$ & \\ 
\rowcolor{white} \multirow{-2}{*}{Vladislavleva-3} & \multirow{-2}{*}{2} & \multirow{-2}{*}{$e^{-x_1}*x_1^3*(cos x_1*sin x_1)*(cos x_1*sin^2 x_1-1)*(x_2-5)$} & $x_2: E[0.05, 10.05, 2]$ & $x_2: E[-0.5, 10.5, 0.5]$ & \multirow{-2}{*}{\cite{miranda2017how}}\\

  \rowcolor{gray!25}            Vladislavleva-4 & 5 & $\frac{10}{5+\sum_{i=1}^5 (x_i-3)^2}$ & $U[0.05, 6.05,1024]$ & $U[-0.25, 6.35, 5000]$ & \cite{lacava2015genetic,medernach2016new,nicolau2016managing,lacava2016epsilon,oliveira2016dispersion}\\ 

\rowcolor{white} &  &  & $x_1: U[0.05, 2,300]$ & $x_1: E[-0.05, 2.1, 0.15]$ & \\ 
& & & $x_2: U[1, 2,300]$ & $x_2: E[0.95, 2.05, 0.1]$ \\
\rowcolor{white} \multirow{-3}{*}{Vladislavleva-5} & \multirow{-3}{*}{3} & \multirow{-3}{*}{$30*(x_1-1)*\frac{x_3-1}{(x_1-10)*x_2^2}$} & $x_3: U[0.05, 2,300]$ & $x_3: E[-0.05, 2.1, 0.15]$ & \multirow{-3}{*}{\cite{chen2016improving,thuong2017combining}} \\

\rowcolor{gray!25}          Vladislavleva-6 & 2 & $6*sin(x_1)*cos(x_2)$ & $U[0.1, 5.9, 30]$ & $E[-0.05, 6.05, 0.02]$ & \cite{chen2016improving,thuong2017combining}\\ 
\rowcolor{white}            Vladislavleva-7 & 2 & $(x_1-3)*(x_2-3)+2*sin((x_1-4)*(x_2-4))$ & $U[0.05, 6.05,300]$ & $U[-0.25, 6.35,1000]$ & \cite{miranda2017how}\\
\rowcolor{gray!25}          Vladislavleva-8 & 2 & $\frac{(x_1-3)^4+(x_2-3)^3-(x_2-3)}{(x_2-2)^4+10}$ & $U[0.05, 6.05,50]$ & $E[-0.25, 6.35, 0.2]$ & \cite{chen2016improving,thuong2017combining}\\
            \midrule[\heavyrulewidth] 
        \end{tabular}
    \end{center}
    \textsuperscript{1} Note that the domain may include values for which the function is undefined.\\
    (*) Datasets not recommended as benchmarks in \cite{white2013better}.\\
    \label{table:synth_datasets}
\end{table*}
    
\end{document}